\documentclass{sna}%
\usepackage{amsmath}
\usepackage{amsfonts}
\usepackage{amssymb}
\usepackage{graphicx}
\usepackage{hyperref}%
\providecommand{\U}[1]{\protect\rule{.1in}{.1in}}
\begin{document}
%
\info{Building a Fuel Moisture Model for\\\vspace*{-3pt}%
the Coupled Fire-Atmosphere Model WRF-SFIRE from Data:\\From
Kalman Filters to Recurrent Neural Networks}
{J.~Mandel\,${}^1$, J.~Hirschi\,${}^1$, A.~K.~Kochanski\,${}%
^2$, A.~Farguell\,${}^2$,
J.~Haley\,${}^3$, D.~V.~Mallia\,${}^4$, B.~Shaddy\,${}^5$, A.~A.~Oberai\,${}%
^5$, and K.~A.~Hilburn\,${}^3$}
{${}^1$University of Colorado Denver, Denver, CO\\
${}^2$San Jos\'e State University, San Jos\'e, CA\\
${}^3$Colorado State University, Fort Collins, CO\\
${}^4$University of Utah, Salt Lake City, UT\\
${}^5$University of Southern California, Los Angeles, CA\\}
\vspace*{-0.3in}
\begin{center}
\emph{Dedicated to the memory of Professor Radim Blaheta}
\end{center}
\vspace*{-0.3in}%

\section{Introduction}

The WRF-SFIRE~modeling system \cite{Mandel-2014-RAA,Mandel-2019-IDH} couples
Weather Research Forecasting (WRF) model with a wildfire spread model and a
fuel moisture content (FMC) model. The FMC is an important factor in wildfire
behavior, as it underlies the diurnal variability and different severity of
wildfires. The FMC model uses atmospheric variables (temperature, relative
humidity, rain) from Real-Time Mesoscale Analysis (RTMA) to compute the
equilibrium FMC and then runs a simple time-lag differential equation model of
the time evolution of the FMC. In the \emph{learning phase}, the model
assimilates \cite{Vejmelka-2016-DAD} FMC observations from sensors on Remote
Automated Weather Stations (RAWS) \cite{NWCG-2019-NSF}, using the augmented
extended Kalman filter. In the \emph{forecast phase}, the model runs from the
atmospheric state provided by WRF without the Kalman filter, since the sensor
data are still in future and not known (Fig.~\ref{fig:kf}).

We seek to improve the accuracy of both the FMC model and of the data
assimilation. The time-lag model represents the FMC in a wood stick by a
single number, while more accurate models use multiple layers
\cite{vanderKamp-2017-MFS} or a continous radial profile
\cite{Nelson-2000-PDC}. Also, the Kalman filter assumes Gaussian probability
distributions and a linear model, while more sophisticated data assimilation
methods can represent more general distributions and allow nonlinear models.
It is, however, unclear how much additional sophistication is worthwhile given
the available data. Thus, we want to build a~model together with data
assimilation directly from data instead. We propose to use a Recurrent Neural
Network\ (RNN) for this.

\section{The FMC model with Kalman filter}

We briefly describe the model from \cite{Mandel-2014-RAA} with data
assimilation from \cite{Vejmelka-2016-DAD}. For simplicity, we consider here
only the situation at a single RAWS location, without rain, and with a single
fuel class with 10h time lag. See \cite{Mandel-2014-RAA,Vejmelka-2016-DAD} for
details, references, and a more general case.%

\begin{figure}[tb]
\begin{center}
\includegraphics[width=6in]{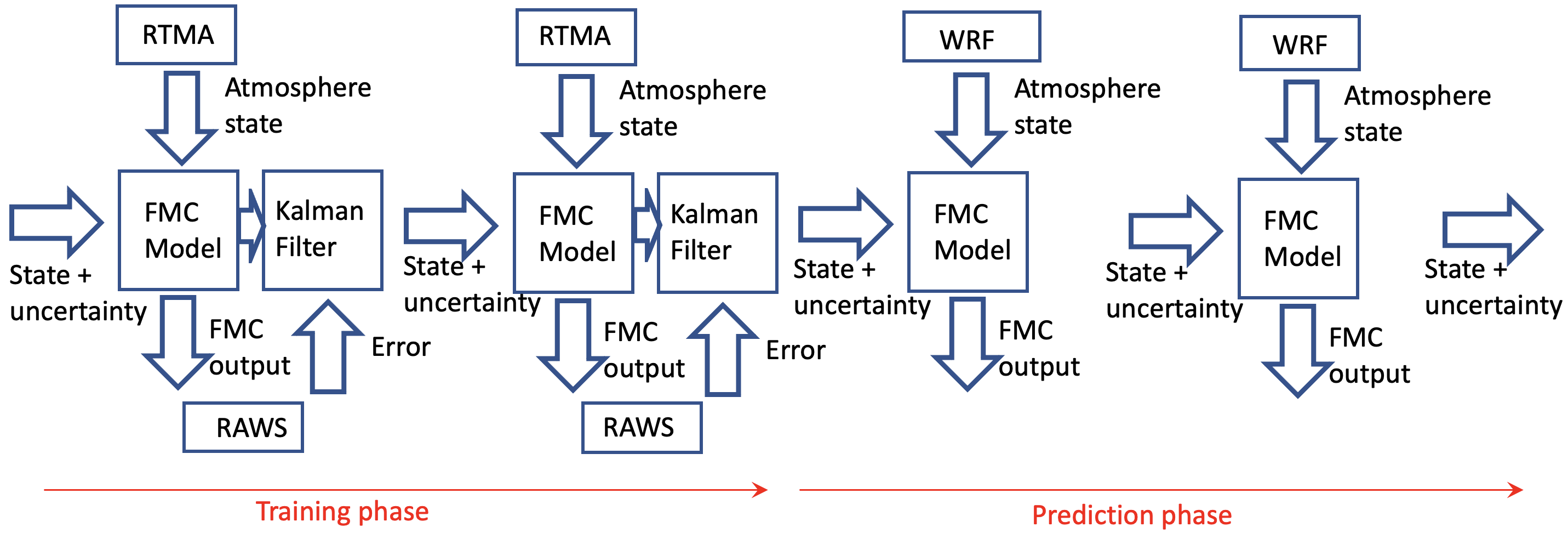}
\vspace*{-0.1in}
\caption{Data flow in the existing model with Kalman filter.}
\label{fig:kf}
\end{center}
\end{figure}%
\begin{figure}[tb]
\begin{center}
\includegraphics[width=6.2in]{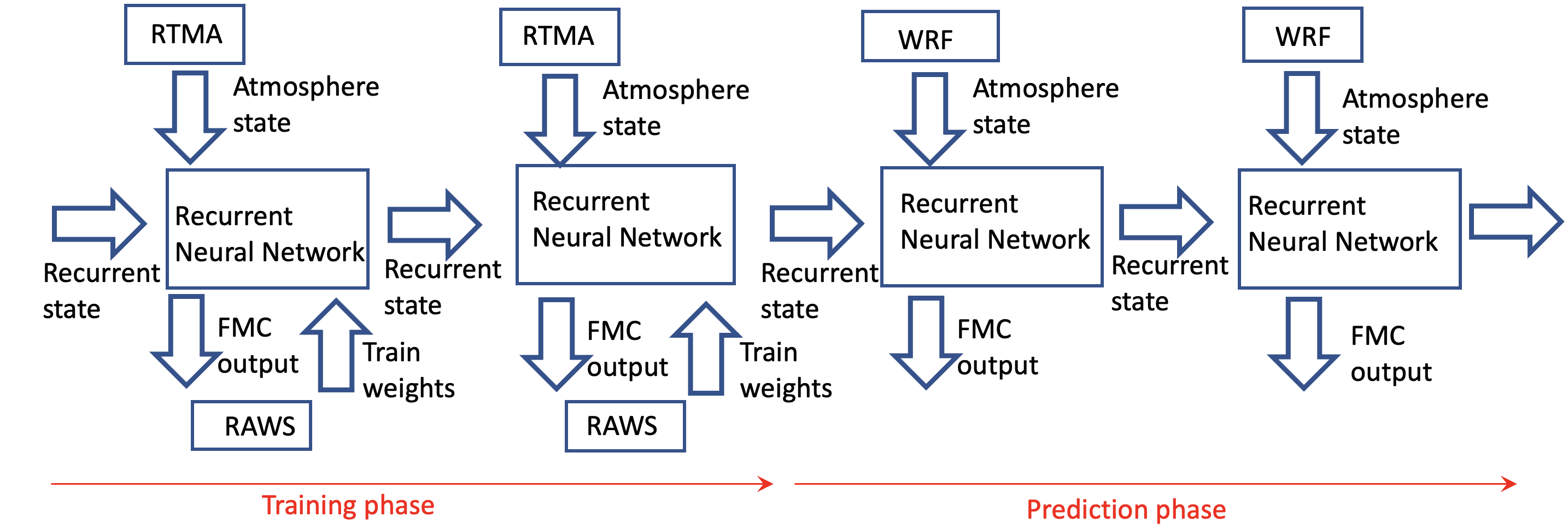}
\vspace*{-0.1in}
\caption{Data flow with the Recurrent Neural Network.}
\label{fig:rnn}
\end{center}
\end{figure}%

The FMC $m(t)$ in wood is the mass of water as \% of the mass of dry wood, and
it changes with time $t$ and atmospheric conditions. A simple empirical model
of the evolution of $m(t)$ in a wood stick in constant atmospheric conditions
is the stick losing water if $m\left(  t\right)  >E_{d}$, the drying
equilibrium, and gaining water if $m\left(  t\right)  <E_{w}$, the wetting
equilbrium, with a characteristic time constant $T$ given by the stick
diameter ($T=10$h for 10h fuel). The values of $E_{w}$ and $E_{d}$,
$E_{w}<E_{d}$, are computed from atmospheric conditions, namely relative
humidity and temperature. We add to both a correction $\Delta E$, assumed
constant in time and to be identified from data. This gives a system of
differential equation on the interval $\left[  t_{k},t_{k+1}\right]  $ for the
augmented state $u=\left(  m,\Delta E\right)  $ of dimension 2,
\begin{align*}
\frac{dm}{dt} &  =\frac{E_{w}+\Delta E-m(t)}{T}\text{\ if }m(t_{k}%
)<E_{w}+\Delta E,\quad\frac{dm}{dt}=\frac{E_{d}+\Delta E-m(t)}{T}\text{\ if
}m(t_{k})>E_{d}+\Delta E,\quad\\
\frac{dm}{dt} &  =0\text{ if }E_{w}+\Delta E\leq m(t_{k})\leq E_{d}+\Delta
E,\quad\frac{d\Delta E}{dt}=0.
\end{align*}
We apply the extended Kalman filter to the evolution $u\left(  t_{k}\right)
\mapsto u\left(  t_{k+1}\right)  $ with the observations $m\left(
t_{k}\right)  =d\left(  t_{k}\right)  $+noise.

The FMC\ Model and\ Kalman filter in the data flow in Fig.~\ref{fig:kf}
implement a nonlinear operator.
Since the operator is the same at all times $t_{k}$, it is applied
recurrently. We seek to identify this operator by a Neural Network (NN)
(Fig.~\ref{fig:rnn}), which then becomes a RNN.

\section{Recurrent Neural Network}

Filtering as an application of NNs is now classical. E.g., RNN was trained to
match the Kalman filter~\cite{DeCruyenaere-1992-CBK}
and synthetizing neural filters \cite{Lo-1994-SAO} estimate both an optimal
filter and a model.
For contemporary RNN\ basics, see, e.g., \cite[Ch.~15]{Geron-2019-HML}. Our
goal is to build a RNN in the context of current high-performance high-level
software, such as Keras, to translate a time series of the atmospheric data in
the form of features $E_{d}$, $E_{w}$ to a~time series of FMC values $m$.%

\begin{figure}[tb]
\begin{center}
\includegraphics[width=6in]{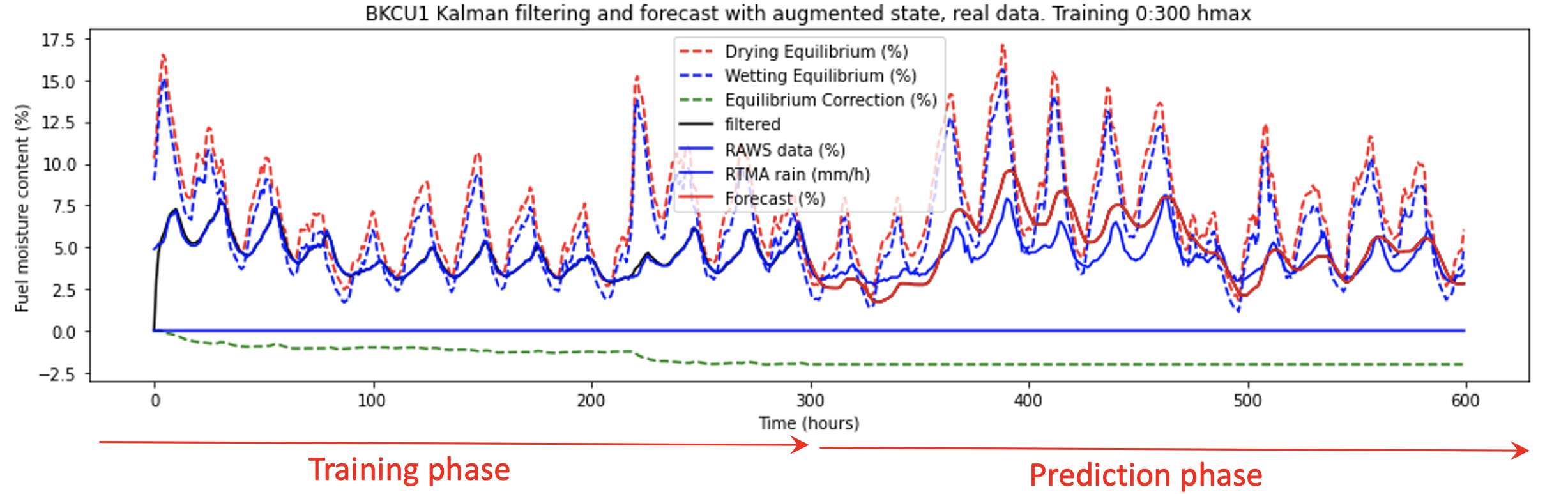}
\vspace*{-0.1in}
\caption
{Learning and forecast with the time-lag model and Kalman filter. The equilibrium correction $\Delta
E$
stabilizes in the training.
Note a large prediction error from 300 to 600 hours.}
\label{fig:kf-result}`
\end{center}
\end{figure}%
%

\begin{figure}[tb]
\vspace*{-0.2in}
\begin{center}
\includegraphics[width=6in]{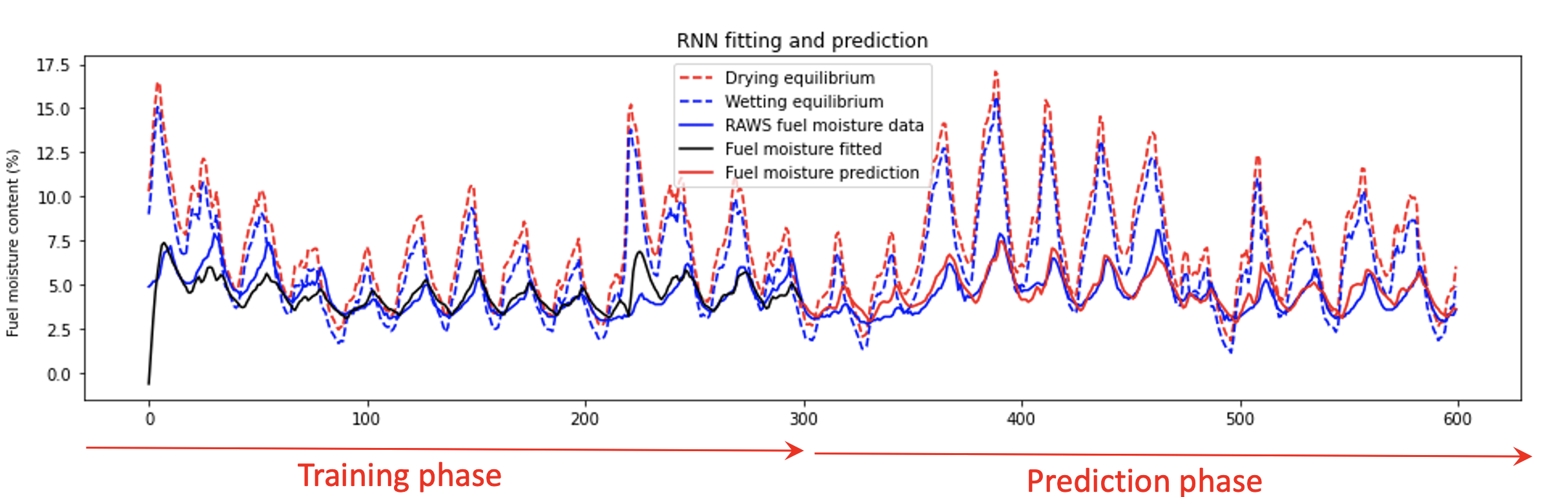}
\vspace*{-0.1in}
\caption{Training and prediction with the Recurrent Neural Network.}
\label{fig:rnn-result}
\end{center}
\end{figure}%

Training RNNs is known to be tricky. One reason is that computing the gradient
of the loss function by back propagation uses the chain rule applied to the
NN\ operator composed with itself many times, which results in
\textquotedblleft vanishing\textquotedblright\ or \textquotedblleft
exploding\textquotedblright\ gradients. To overcome this, we train a stateful
RNN model \cite[p.~532]{Geron-2019-HML} and limit the number of times the
NN\ operator is composed with itself to a small number of $s$ timesteps. In
each batch, the built-in stochastic gradient (SG) optimizer in Keras is
presented with a sequence of training samples, each of the form of a short
sequence of (input$_{k+1}$,\ldots,input$_{k+s}$) and (target$_{k+s}$),
$k=1,2,\ldots$. The inputs are the features $\left(  E_{d},E_{w}\right)  ,$
and the targets are the observations from the RAWS FMC sensors. The
NN\ operator is applied to the recurrent state (hidden$_{k+1}$,\ldots
,hidden$_{k+s}$) and the input to produce the new recurrent state
(hidden$_{k+2}$,\ldots,hidden$_{k+s+1}$) and (output$_{k+s}$) which is
compared with (target$_{k+s}$) to compute a contribution to the loss function
and its gradient. After the RNN is trained, the optimized weights are copied
to an identical stateless NN model \cite[p.~534]{Geron-2019-HML}, which is
then used for the evaluation of the NN\ operator in the prediction phase.

Though this procedure is commonly used, it did not work well in this
application and the resulting forecast was much worse than when using the
extended KF. However, it is straightforward to implement a version of the
Euler method for the time-lag differential equation $dm/dt=\left(  E-m\right)
/T$ by a single neuron with linear activation and a suitable choice of
weigths,
\[
m_{k+1}=e^{-\Delta t/T}m_{k}+\left(  1-e^{-\Delta t/T}\right)  E_{k}%
,\qquad\Delta t=t_{k+1}-t_{k}.
\]
Here, $m_{k}$ is the hidden state, also copied to the output, and $E_{k}$ is
the input. The single neuron RNN worked well on synthetic examples, so we used
a hidden layer with linear activation, pre-trained by using the initial
weights above.

Our final network had one hidden layer of $6$ neurons pre-trained as above,
and two input neurons and one output, all with linear activation. We chose the
dimension of the hidden state $6$ to accommodate different time scales. The
training used $s=5$ timesteps. The resulting prediction
(Fig.~\ref{fig:rnn-result}) was better than from the differential equation
model with extended KF (Fig.~\ref{fig:kf-result}).


\section{Conclusion}

We have used batch training of a stateful RNN\ with linear activation and
initial weights chosen to make the RNN an exact model in a special case. A
hidden layer of several neurons initialized to the same weights then produced
a better prediction than a differential equation model with extended KF.
Exploiting this principle with more general activations, such as RELU, may
enable switching between different behaviours, such as drying, wetting, or
rain, or quantification of uncertainty in future.

\textbf{Acknowledgement:} This work was partially supported by NASA grants
80NSSC19K1091, 80NSSC22K1717, and 80NSSC22K1405.

\bibliographystyle{siam}
\bibliography{../../references/ml,../../references/bddc,../../references/geo,../../references/other,../../references/slides}

\end{document}